\documentclass{article} 
\usepackage{nips15submit_e,times}
\usepackage{hyperref}
\usepackage{amsmath}
\usepackage{graphicx}
\usepackage{multicol}
\usepackage{url}

\title{Feedforward Sequential Memory Neural Networks without Recurrent Feedback}
\author{Shiliang Zhang$^1$, Hui Jiang$^2$, Si Wei$^3$, Lirong Dai$^1$ \\
	$^1$National Engineering Laboratory for Speech and Language Information Processing \\
	University of Science and Technology of China, Hefei, Anhui,  China\\
	$^2$Department of Electrical Engineering and Computer Science \\
	York University,  4700 Keele Street, Toronto, Ontario, M3J 1P3, Canada\\
	$^3$IFLYTEK Research, Hefei, Anhui,  China\\
	{\tt \small zsl2008@mail.ustc.edu.cn, hj@cse.yorku.ca, siwei@iflytek.com, lrdai@ustc.edu.cn}
}

\nipsfinalcopy 

\begin{document}

\maketitle

\begin{abstract}
We introduce a new structure for memory neural networks, called feedforward sequential memory networks (FSMN), which can learn long-term dependency without using recurrent feedback. The proposed FSMN is a standard feedforward neural networks equipped with learnable sequential memory blocks in the hidden layers. In this work, we have applied FSMN to several language modeling (LM) tasks. Experimental results have shown that the memory blocks in FSMN can learn effective representations of long history. Experiments have shown that FSMN based language models can significantly outperform not only feedforward neural network (FNN) based LMs but also the popular recurrent neural network (RNN) LMs.
\end{abstract}

\section{Introduction} 
When machine learning methods are applied to model sequential data such as text, speech and video, it is very important to take advantage of the long-term dependency. Traditional approaches have explored to capture the long-term structure within the sequential data using recurrent feedbacks such as in regular recurrent neural networks (RNNs) or LSTM-based models.  \cite{LSTM1997,Graves2013,Chung2014,Mikolov2014}. RNNs can learn and carry out complicated transformations of data over extended periods of time and store the memory in the weights of the network. Therefore, they are gaining more and more popular in sequential data modeling tasks. More recently, different from RNNs, there has also been a surge in constructing neural computing models with varying forms of explicit memory units \cite{Graves2014NTM,Weston2014MN,Sukhbaatar2015MN,Joulin2015StackRNN} . For example, in \cite{Weston2014MN}, the proposed memory networks employ a memory component that can be read from and written to. In \cite{Graves2014NTM}, the proposed neural turing machines (NTM) improve the memory of neural networks by coupling with external memory resources, which can learn to sort a small set of numbers as well as other symbolic manipulation tasks.

In this work, we have proposed a simpler structure for memory neural networks, namely feedforward sequential memory networks (FSMN), which can learn long-term dependency in sequential data without using the recurrent feedback.  For FSMN, we extend the standard feedforward neural networks by introducing memory blocks in the hidden layers. Different from RNNs, the overall FSMNs remain as a feed-forward structure so that they can be learned in much more efficient and stable ways than RNNs and LSTMs.  In our work, we have evaluated the performance of FSMN on two language modeling tasks: Penn Tree Bank (PTB) and English wiki in Large Text Compression Benchmark (LTCB). In both tasks, FSMN based language models can significantly outperform not only the standard FNN-LMs but also the popular recurrent neural network (RNN) LMs by a significant margin.

\section{Our Approach}
\label{sec.FSMN}
\subsection{Feedforward Sequential Memory Neural Networks}
\begin{figure}
\centering
\includegraphics[width=0.9\linewidth]{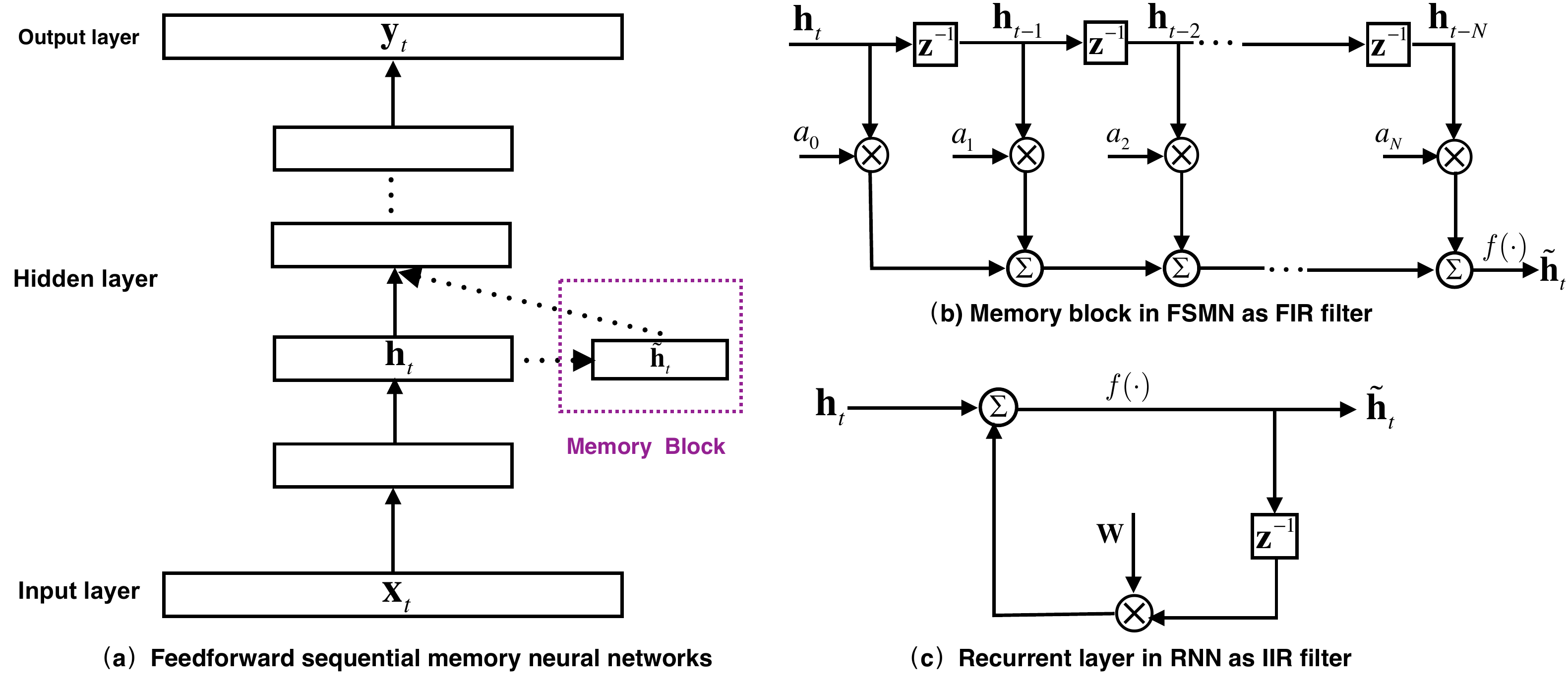}
\caption{Illustration of feedforward sequential memory networks and comparison with RNNs.}
\label{fig:FSMN}
\end{figure}
Feedforward sequential memory network (FSMN) is a standard feedforward neural network with single or multiple memory blocks in the hidden layer. For instance,  Figure \ref{fig:FSMN} (a)  shows an FSMN with one  memory block added into its second hidden layer.  Given a sequence, $\bf{X} = \{\bf{x}_1, \bf{x}_2 , \cdots,   \bf{x}_T \}$,  each $\bf{x}_t \in \mathcal{R}^{D\times 1}$ represents an input data at time instance t. The corresponding hidden layer outputs are denoted as $\bf{H} = \{\bf{h}_1, \bf{h}_2 , \cdots,  \bf{h}_T \}$, each $\bf{h}_t \in \mathcal{R}^{D_\ell\times 1}$.  As shown in Figure \ref{fig:FSMN} (b), we can use a tapped-delay structure to encode $\bf{h}_t$ and its previous $N$ histories into a fixed-sized representation in the memory block:
\begin{equation}\label{eq.1}
\mathbf{\tilde h}_t= f(\sum\limits_{i = 0}^N a_i \cdot \mathbf{h}_{t-i})
\end{equation}
where all coefficients form an N-dimension learnable vector ${\bf a}=\{a_0, a_1, a_2,\cdots, a_N\}$, and $f(\cdot)$ is the activation function (sigmoid or RELU). Furthermore, as shown in 
Figure \ref{fig:FSMN} (a), $\bf{\tilde h}_t$ may be fed into next hidden layer in the same way as $\bf{h}_t$. 
 
From the viewpoint of signal processing,  the memory block in FSMN can be regarded as a high-order finite impulse response (FIR) filter while the recurrent layer in RNNs, namely 
$\mathbf{\tilde h}_t=f(\mathbf{h}_{t}+\bf{W}\cdot\mathbf{\tilde h}_{t-1})$,
may be viewed as a first-order infinite impulse response (IIR) filter, see Figure \ref{fig:FSMN} (c).  Obviously, the vector $\bf{a}$ may  be regarded as the coefficients of an N-order FIR filter. We know IIR filters are more compact than FIR filters. However, IIR filters may be difficult to implement.
In some cases, IIR filters may become unstable but FIR filters are always stable.
Moreover, the learning of IIR-filter-like RNNs requires to use the so-called back-propagation through time (BPTT) which significantly increases the computational complexity of the learning and  also causes the problems of gradient vanishing and exploding \cite{Bengio1994}. On the other hand, the proposed FIR-filter-like FSMNs can be efficiently learned using the standard back-propagation procedure.
Therefore, the learning of FSMNs is more stable and efficient than that of RNNs.

\subsection{Implement FSMN for language models}

The goal in language modeling is to predict the next word in a text sequence given all previous words. We now explain how to implement FSMNs for this task. FSMN is a standard feedforward neural network (FNN) except the additional memory blocks. We will show that the memory block can be efficiently implemented as sentence-by-sentence matrix multiplications, which are suitable for the mini-batch based stochastic gradient descent (SGD) method running on GPUs. 

Suppose the N-order FIR filter coefficients in the memory block are denoted as ${\bf a} =\{a_0, a_1, a_2,\cdots, a_N\}$. For a given sentence $\bf{X}$ consisting of $T$ words, we may construct a $T \times T$ square matrix $M$ as follow:
\begin{equation}\nonumber
{\mathbf{M}} = \left[ \begin{gathered}
  {a_0}\;\;{a_1} \cdots \;{a_N}\quad 0\; \cdots \;0 \cdots  \hfill \\
  0\quad {a_0}\quad {a_1} \cdots \;{a_N}\;\;\;0 \cdots  \hfill \\
   \vdots \;\; \cdots \;0\quad  \ddots \quad \quad  \ddots \;\; \vdots  \hfill \\
  0\quad  \cdots \quad \qquad {a_0}\; \cdots \;\;{a_N} \hfill \\
   \vdots \qquad  \cdots \quad \qquad \quad \ddots \;\;\; \vdots  \hfill \\
  0\quad  \cdots \qquad \qquad \;\; 0\quad\; {a_0} \hfill \\ 
\end{gathered}  \right]
\end{equation}
Therefore, the sequential memory operations in eq.(\ref{eq.1}) for the whole sequence can be computed with one matrix multiplication as: $\mathbf{\tilde H} = f({\mathbf{H}} \cdot {\mathbf{M}})$.
 
Similarly, we may extend the idea to a mini-batch composed of $K$ sentences, ${\cal L}=\{ \bf{X}_1 \;  \bf{X}_2 \cdots \bf{X}_K\}$, we can compute the sequential memory representation for all sentences in the mini-batch as follows:
\begin{equation}
{\bf \tilde{H}} =f(\left[\bf{H}_1, \bf{H}_2\cdots \bf{H}_K\right]\cdot\left[ \begin{gathered}
{\bf M}_1 \hfill \\
\qquad  {\bf M}_2 \hfill \\
\qquad \qquad  \ddots  \hfill \\
\qquad \qquad \qquad  {\bf M}_K\hfill \\ 
\end{gathered}  \right] )= f({\bf \bar{H}}\cdot {\bf \bar{M}})
\end{equation}

In the backward pass, except the weights in the network, we also need to calculate the gradients of ${\bf \bar{M}}$, which is then used to update the filter coefficients $\bf{a}$. We can calculate the gradients using the standard back-propagation (BP) algorithm. Therefore,  all computation in FSMNs can be formulated as large matrix multiplications, which can be efficiently conducted in GPUs. As a result, FSMN based LMs 
have the same computational complexity as the standard NN LMs in training, which is much more efficient than RNN-LMs.

\section{Experiments}

We have evaluated FSMNs on two benchmark LM tasks: i) the Penn Treebank (PTB) corpus of about 1M words, following the same setup as \cite{Mikolov2011Extension}. 
ii) The Large Text Compression Benchmark (LTCB) \cite{Mahoney2011}. In LTCB, we use the {\em enwik9} dataset, which is composed of  the first $10^9$ bytes of enwiki-20060303-pages-articles.xml. We split it into three parts:  training (153M), validation (8.9M) and test (8.9M) sets. 
We limit the vocabulary size to 80k for LTCB and replace all out-of-vocabulary words by $<$UNK$>$.

For FSMNs, the hidden units employ the rectified linear activation function, i.e., $f(x)=\max(0,x)$. The nets are initialized based on the normalized initialization in \cite{Glorot2010}, without using any pre-training.  We use SGD with a mini-batch size of 200 and 500 for PTB and LTCB tasks respectively. The initial learning rate is 0.4 and 0.002 for the weights and filter coefficients respectively, which is kept fixed as long as the perplexity on the validation set decreases by at least 1. After that, we continue six more epochs of training, where the learning rate is halved after each epoch. In PTB task, we also use momentum (0.9) and weight decay (0.00004) to avoid overfitting.

\subsection{Results}

We have first evaluated the performance of FSMN-LMs on the PTB task. We have trained FSMN with an input context window of two, where the previous two words are used to predict the next word. The FSMN contains a linear projection layer (of 200 nodes), two hidden layers (of 400 nodes pre layer) and a memory block in the first hidden layer. For the memory block, we use a 20-order FIR filter. In Table 1, we have summarized the perplexities on the PTB test set for various language models.

For the LTCB task, we have trained several  baseline systems: i) two n-gram LMs (3-gram and 5-gram) using the modified Kneser-Ney smoothing without count cutoffs; ii) several traditional feedforward NNLMs with different model sizes and input context windows (bigram, trigram, 4-gram and 5-gram); iii) an RNN-LM with one hidden layer of 600 nodes using the toolkit in \cite{Mikolov2010recurrent}; iv) 2nd-order FOFE based FNNLM \cite{Zhang2015} with different hidden layer sizes. Moreover, we have examined our FSMN based LMs with different architectures.
We have trained a 3-hidden-layer FSMN with a memory block in the first hidden layer, second hidden layer or both.  The order of the FIR filter is 30 in these experiments. In Table 2, we have summarized the perplexities on the LTCB test set for various models. 

Experimental results in Table 1 and Table 2 have shown that the FSMN based LMs can significantly outperform the baseline higher-order feedforward neural network (FNN) LMs, FOFE-based FNN LMs as well as the popular RNN-based LMs by a quite significant margin.

\begin{table}[t]
	\begin{minipage}{0.35\linewidth}
	\begin{small}
		\caption{Perplexities on PTB for various LMs.}
		\begin{tabular}{|l|c|}
			\hline 
			Model  & Test PPL \\\hline 
			KN 5-gram \cite{Mikolov2011Extension}      &  141 \\
			RNNLM \cite{Mikolov2011Extension}  & 123 \\
			LSTM \cite{Graves2013}   &  117   \\ 
			MemN2N\cite{Sukhbaatar2015MN} & 111\\
			trigram FNNLM\cite{Zhang2015} & 131 \\
			6-gram FNNLM\cite{Zhang2015} & 113 \\ 
			FOFE-FNNLM\cite{Zhang2015} & 108 \\ \hline
			FSMN-LM                                 & \textbf{102}\\\hline
		\end{tabular} 
	\end{small}
	\label{tab:PTB_summary}
\end{minipage}
\hfill
	\begin{minipage}{0.6\linewidth}
	\begin{small}
	\caption{Perplexities on LTCB for various language models. (M) denotes a hidden layer with memory block.} 
	\begin{tabular}{|l|l|c|}
		\hline
		Model & Architecture & Test PPL \\\hline
		KN 3-gram &   -                &    156 \\
		KN 5-gram &  -                 &   132  \\\hline
		RNN-LM   & [1*600]-80k   &   112 \\\hline
		FNNLM     & [2*200]-3*600-80k & 155 \\                
		                 & [2*200]-1200-2*600-80k & 154 \\\hline
		FOFE-FNNLM & [2*200]-3*600-80k &  104\\
	                	 & [2*200]-1200-2*600-80k & 107 \\\hline
	    FSMN-LM & [2*200]-600(M)-600-600-80k & 95\\
	                     &  [2*200]-600-600(M)-600-80k & 96\\
	                     &  [2*200]-600(M)-600(M)-600-80k & \textbf{92}\\\hline
	\end{tabular} 
	\end{small}
	\label{tab:WIKI_summary}
\end{minipage}
\end{table}

\section{Conclusions and Future Work}
In this work, we have proposed a novel neural network architecture, namely feedforward sequential memory networks (FSMN), which use FIR-filter-like memory blocks in the hidden layer of standard feedforward neural networks. Experimental results on language modeling tasks have shown that the FSMN can effectively learn the long term history. For the future work, we will try to apply this model to other sequential data modeling tasks, such as acoustic modeling in speech recognition.

%
\end{document}